\documentclass[11pt]{article}
\pdfoutput=1
\usepackage[margin=1in]{geometry}
\usepackage{mathtools}
\usepackage{amssymb,amsthm}
\usepackage{algorithmic,algorithm}
\usepackage{subfigure}
\usepackage{caption}
\usepackage{verbatim,url}
\usepackage{natbib} %,natbibspacing
\usepackage{authblk}

\newcommand{\mb}[1]{\mathbf{#1}}

\title{Inference in Kingman's Coalescent with Particle Markov Chain Monte Carlo Method}

\author[1]{Yifei Chen}
\author[1,2,*]{Xiaohui Xie}

\affil[1]{Department of Computer Science, University of California Irvine}
\affil[2]{Institute for Genomics and Bioinformatics, University of California Irvine}
\affil[*]{\emph{email}: \texttt {xhx@ics.uci.edu}}

\begin{document}

\maketitle

\begin{abstract}
We propose a new algorithm to do posterior sampling of Kingman's coalescent, based upon the Particle Markov Chain Monte Carlo methodology. Specifically, the algorithm is an instantiation of the Particle Gibbs Sampling method, which alternately samples coalescent times conditioned on coalescent tree structures, and tree structures conditioned on coalescent times via the conditional Sequential Monte Carlo procedure. We implement our algorithm as a C++ package, and demonstrate its utility via a parameter estimation task in population genetics on both single- and multiple-locus data. The experiment results show that the proposed algorithm performs comparable to or better than several well-developed methods.
\end{abstract}

%The produces high quality samples and converges fast, because the cSMC particles can represent the posterior distribution well.

\section{Introduction}
Data shows hierarchical structure in many domains. For example, computer vision problems often involve hierarchical representation of images \citep{lee2009convolutional}. In text mining, documents can be modeled as hierarchical generative processes \citep{blei2003latent, teh2006hierarchical}. Algorithms that can effectively deal with hierarchical structure play an important role in uncovering the intrinsic structures of data.

We focuses on a hierarchical model in population genetics, the Kingman's n-colescent \citep{kingman1982coalescent, kingman1982genealogy}, shortly referred to as the coalescent. It is a tree-structured process that models the genealogical relationship among a set of individuals, represented by some feature vectors (typically the DNA sequences). Although forward simulation of genealogy tree \citep{stephens2000inference} is simple, posterior inference is much difficult due to the complicate state space, which requires combinatorial search over tree structures and high-dimensional sampling of coalescent time series. Many works have been done in this area, and are roughly grouped into three methodologies, i.e., Importance Sampling (IS),  Markov Chain Monte Carlo (MCMC), and Sequential Monte Carlo (SMC). The IS approaches \citep{griffiths1994ancestral, griffiths1994simulating, stephens2000inference} simulate mutation events explicitly, and require coalescent events happen between identical sequences, so aren't efficient for data of highly diverse sequences. The MCMC approach \citep{kuhner1995estimating, kuhner2006lamarc} uses random structure modification to explore the tree space. Although the samples are cheap to simulate computationally, they don't represent the posterior distribution well, which makes the whole algorithm not accurate. The SMC approaches \citep{teh2008bayesian, gšrŸr2009efficient} use the so-called ``local posterior" to propose coalescent events \& times. The pair-wise numerical integration may bring potential scalability problem. But recent improvement using ``pair similarity" heuristic seems to alleviate the problem in some sense \citep{gšrŸrscalable}.

In this paper, we propose a novel inference algorithm for the coalescent based upon the Particle Markov Chain Monte Carlo (PMCMC), a methodology recently developed in statistics community \citep{andrieu2010particle}. More specifically, it falls under the framework of the Particle Gibbs Sampling (PGS), which is a case of PMCMC that targets joint posterior distribution of parameter and hidden variable. Our core idea is to decouple coalescent times (tree branching lengths) and tree structure, view th e former as the parameter while the later as the hidden variable, and alternately sample one conditioned on another. Specifically, we simulate the coalescent times by Gibbs sampling, and simulate the tree structure by the conditional Sequential Monte Carlo (cSMC). Although computationally expensive, this approach explores the state space and generates tree structures informatively.  We demonstrate this by estimating the likelihood surface of $\theta$, a parameter of the coalescent which captures the mutation rate and the size of the background population. Experiments show that our method performs comparable to or better than the well-developed methods as mentioned previously.

The following paper is organized as follows: In Section 2, we introduce the coalescent in the context of graphical models and probabilistic inference. In Section 3, we explain the challenging problems of coalescent posterior sampling and genetic parameter estimation, and review current approaches. In Section 4, we formally describe our algorithm, Particle Gibbs Sampler for the Kingman's Coalescent (PGS-Coalescent). Experimental results are presented and analysed in Section 5. We give more discussion and conclude in Section 6.

\section{Kingman's n-Coalescent}

\begin{figure}[t]
%\vspace{1in}
\center
\includegraphics[width=2.7in]{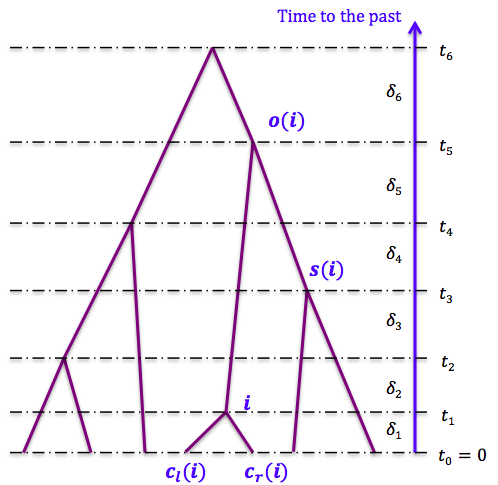}
\caption{An Example of Coalescent Process}
\label{Fig:Genealogy}
\end{figure}

Kingman's coalescent is proposed in population genetics to model the genealogy upon the haploid population \citep{kingman1982coalescent, kingman1982genealogy}. Briefly speaking, the coalescent is a tree-structured process that starts from the current population, and evolves backward with random mutation and coalescent events until the Most Recent Common Ancestor (MRCA) is reached. Besides the successful applications in population genetics, it also becomes popular as a hierarchical clustering model in machine learning \citep{teh2008bayesian,gšrŸr2009efficient, gšrŸrscalable}. The state space representation of the coalescent varies among different inference algorithms and applications, but falls into two broad categories. One represents the coalescent events and the coalescent times but not mutation events or their times \citep{kuhner1995estimating, slatkin2002vectorized, gšrŸrscalable}, while the other represents the coalescent and mutation events but not their times \citep{griffiths1994ancestral, griffiths1994simulating, stephens2000inference}. In this paper we adopt the former representation. Figure \ref{Fig:Genealogy} illustrates the coalescent tree under such representation. Notations are summarized as follows.

Denote a subset of a current population as $X=\{x_1,...,x_n\}$, where $n$ is the number of individuals in the subset. They compose the leaf nodes of the coalescent. Starting from $X$, $n-1$ coalescent events will happen until MRCA is reached. They are represented as the hidden nodes, and fully describe the coalescent structure. Further more, these events happen sequentially at time $\mathcal{T} = \{t_1,...,t_{n-1}\}$, with hidden states $Y=\{y_1,...,y_{n-1}\}$. For any event $i$, if applicable, denote its waiting time since previous event $i-1$ as $\delta_i=t_{i-1}-t_i$, its parent as $o(i)$, its sibling as $s(i)$, and its children as $\{c_l(i), c_r(i)\}$. Mutation events can happen between any pair of parent- and child-nodes, but are represented implicitly as a continuous-time Markov process with transition rate matrix $\theta R$, where $\theta \propto N\mu$ is a genetic parameter that is controlled by the effective population size $N$ and neutral mutation rate per site per generation $\mu$ \citep{kuhner2006lamarc}.  Accordingly, the transition matrix from the parent to the child is $T_{o(i),i}=e^{\theta(t_i-t_{o(i)}) R}$. Finally, denote the equilibrium distribution as $p_0$, which satisfies $p_0^TR=0$.

A coalescent tree $\mathcal{G}$ is specified by the tree structure and the branching lengths.  At each event $i$, any pair of nodes that remain in the population is equal likely to coalesce. The pairing structure of the nodes defines the genealogy structure. The prior probability of any genealogy structure is $ p(\mathcal{S}) = \displaystyle\prod_{i=1}^{n-1} \begin{pmatrix} n-i+1\\2\end{pmatrix}^{-1}$,  while the waiting time $\delta_i$ follows exponential distribution with rate $\begin{pmatrix}n-i+1\\2\end{pmatrix}$. So the prior probability of the coalescent tree $\mathcal{G}=(\mathcal{S},\mathcal{T})$ is,
\begin{align}
\label{coalescent:prior}
p(\mathcal{G}) = p(\mathcal{S})p(\mathcal{T}) = \prod_{i=1}^{n-1} \exp\left\{ -\begin{pmatrix}n-i+1\\2\end{pmatrix}\delta_i \right\}
\end{align}

 \subsection{Belief Propagation in the Coalescent}
If $\mathcal{G}$ and $\theta$ are known, one can solve $p_\theta(X|\mathcal{G})$ elegantly with belief propagation (BP) \citep{felsenstein1981evolutionary, pearl1986fusion}. Note that \citep{teh2008bayesian, gšrŸr2009efficient, gšrŸrscalable} apply the upward part of BP.  The key difference of our algorithm is to do propagation on both directions, which allows Gibbs Sampling targeting $p_\theta(\mathcal{T}|X,\mathcal{S})$.  While algorithmic details will be discussed in next section, here we provide a brief description of BP.

BP in coalescent has two rounds. First, the leaf nodes propagate messages upward until the root. Second, the root collects incoming messages and propagates them downward until the leaves. After the two-round propagation, one can do inference at any internal node with incoming messages to it. Formally, the message from a leaf node $i\in\{1,2,...,n\}$ is,
\begin{align}
\label{msg:leaf}
\mb{m}_{i\rightarrow o(i)}^u = \frac{1}{Z_{i\rightarrow o(i)}^u}\delta_{x_i}(x),
\end{align}
the messages associated with a non-root hidden node $i\in\{n+1,n+2,...,2n-2\}$ are,
\begin{align}
\label{msg:internal}
&\mb{m}^u_{i\rightarrow o(i)} = \frac{1}{Z^u_{i\rightarrow o(i)}} \prod_{b=l,r} \sum_{y_{c_b(i)}} T_{i,c_b(i)} \mb{m}_{c_b(i)\rightarrow i}, \nonumber\\
&\mb{m}^d_{i\rightarrow c_b(i)} = \frac{1}{Z^d_{i\rightarrow c_b(i)}} \sum_{y_{o(i)}} T_{o(i),i} \mb{m}_{o(i)\rightarrow i}  \sum_{y_{s(c_b(i))}} T_{i,s(c_b(i))} \mb{m}_{s(c_b(i))\rightarrow i}
\end{align}
and the messages associated to the root node $i=2n-1$ are
\begin{align}
\label{msg:root}
\mb{m}^d_{i\rightarrow c_b(i)} = \frac{1}{Z^d_{i\rightarrow c_b(i)}}  \sum_{y_{s(c_b(i))}} p_0(y_i) T_{i,s(c_b(i))} \mb{m}_{s(c_b(i))\rightarrow i} 			
\end{align}
In Equation \eqref{msg:leaf}, \eqref{msg:internal} and \eqref{msg:root}, $b\in{\{l,r\}}$ represents left and right branches of the selected node. `u' and `d' represent message directions, upward and downward separately. Similar to \citep{teh2008bayesian}, the normalization constant Z is set such that $\sum_v p_0(v) \mb{m}(v)=1$. Hence the likelihood function of $X$ can be expressed at a non-root node as,
\begin{align}
\label{XcG:internal}
p_\theta(X|\mathcal{G}) =  \prod_{j=1, j\neq i}^{2n-2} Z_j^{(i)} \sum_{y_i} \sum_{y_{o(i)}} T_{o(i),i} \mb{m}_{o(i)\rightarrow i}
			     \prod_{b=l,r} \sum_{y_{c_b(i)}} T_{i,c_b(i)} \mb{m}_{c_b(i)\rightarrow i}
\end{align}
and at root node as,
\begin{align}
\label{XcG:root}
p_\theta(X|\mathcal{G}) = \prod_{j=1}^{2n-2} Z_j^{u} \sum_{y_{2n-1}}p_0(y_{2n-1}) \prod_{b=l,r}
			      \sum_{y_{c_b(2n-1)}} T_{2n-1,c_b(2n-1)} \mb{m}_{c_b(2n-1)\rightarrow 2n-1}
\end{align}
where the normalization constants $\{Z_j^{(i)}\}$ are of those messages which direct toward node $i$. The subscript $_\theta$ emphasizes the dependency of the probability to mutation rate through $\{Z_j^{(i)}\}$ and $\{T_{o(i),i}\}$. Accordingly, the joint probability of the data and the coalescent $p_\theta(X,\mathcal{G}) = p_\theta(X|\mathcal{G})p(\mathcal{G})$ can be computed with Equation \eqref{coalescent:prior}, and \eqref{XcG:internal} or \eqref{XcG:root}.

In biological applications $X$ are typically sequence data of multiple sites, for example, allel or DNA sequences. Denote the sequence length as $L$. Assuming each site mutate independently, it is easy to show the message $\mb{m}_{i\rightarrow j}$ and the corresponding normalization constant $Z_{i\rightarrow j}$ are given by
\begin{align}
\mb{m}_{i\rightarrow j} &= \prod_{l=1}^L \mb{m}^{(l)}_{i\rightarrow j} \nonumber\\
Z_{i\rightarrow j} &= \prod_{l=1}^L Z^{(l)}_{i\rightarrow j}
\end{align}
where $\mb{m}^{(l)}_{i\rightarrow j}, Z^{(l)}_{i\rightarrow j}$ are the message and the normalization at each locus $l\in\{1,2,...,L\}$. Therefore, BP still applies to the multiple-site case.

\section{Sampling and Parameter Estimation of the Coalescent}
In real applications, $\mathcal{G}$ and $\theta$ are usually unknown variables which are of interest to geneticists. For example, given a sub-population, people may want to estimate the likelihood surface of $\theta$ \citep{stephens2000inference}, which will shed light on the mutation rate and size of the background population,
\begin{align}
L(\theta) = \int_\mathcal{G} p_\theta(X,\mathcal{G}) \mathrm{d} \mathcal{G}
\end{align}
Exact inference is intractable, because $L(\theta)$ involves combinatorial summation of $n!(n-1)!/2^{n-1}$ possible coalescent structures \citep{stephens2000inference}, each of which involves an $O(n)$-dimensional integration over coalescent times. Therefore, people usually resort to Monte Carlo (MC) methods \citep{liu2008monte}. Researches in this problem roughly belong to three categories, using different methodologies and coalescent representations. In the following, we briefly review the three categories, and a new methodology which will be the basis of our proposed algorithm.

\subsection{Importance Sampling}
\cite{griffiths1994ancestral, griffiths1994simulating} pioneered coalescent inference using \emph{Importance Sampling}. \cite{stephens2000inference} popularized this approach. They represent the coalescent as an ordered list of partitions evolving forward, $\mathcal{G} = {(H_{-m},...,H_{-1},H_0)}$. At each generation, a partition is selected, and then mutation or split is applied to one of its individual. This process goes until population size grows to $n+1$. \cite{stephens2000inference} exploits the character of $p_\theta(\mathcal{G}|X)$, and invents an informative proposal distribution that evolves backward and converge to $p_\theta(\mathcal{G}|X)$ in the parent-independent mutation (PIM) case. In IS, the marginal likelihood of $\theta$ is approximated by,
\begin{align}
\label{lhsurf::IS}
L(\theta) \approx \frac{1}{M} \sum_{i=1}^M \frac{p_\theta(X,\mathcal{G}^{(i)})}{q_{\theta_0}(\mathcal{G}^{(i)}|X)}
\end{align}

\subsection{Markov Chain Monte Carlo}
\citep{kuhner1995estimating} is the first work to apply MCMC to coalescent inference. Its implementation is integrated in the LAMARC software \citep{kuhner2006lamarc}, a very popular analysis tool of population genetic parameters. See \citep{tavarŽ2004part,wakeley2009coalescent} for a review of this work and other MCMC approaches. \citep{kuhner1995estimating} represents coalescent events and time (scaled in units of mutation rate), and uses a Metropolis-Hastings scheme to explore the genealogy space. The proposal distribution is very simple: 1) randomly break a non-root node from its parent. 2) randomly coalesce the node to another branch with exponentially distributed time. Their MCMC approach computes the relative likelihood of $\theta$ as,
\begin{align}
\label{lhsurf::MCMC}
\frac{L(\theta)}{L(\theta_0)} \approx \frac{1}{M} \sum_{i=1}^M \frac{p(\mathcal{G}^{(i)}|\theta)}{p(\mathcal{G}^{(i)}|\theta_0)}
\end{align}

\subsection{Sequential Monte Carlo}
%SMC approaches recently published in machine learning  \citep{teh2008bayesian, gšrŸr2009efficient, gšrŸrscalable} use a similar state space representation. In general, starting from the current population, they iteratively select coalescent pairs and times based upon some distribution called ``local posterior''. Using their notation, this distribution takes the form
%\begin{align}
%\label{PostPost::Proposal}
%q(\delta_i, \rho_{l_i}, \rho_{r_i}) \propto &\exp\left\{-\begin{pmatrix}n-i+1 \\ 2 \end{pmatrix}\delta_i\right\} \nonumber\\
%							      & Z_{\rho_i}(\delta_i, \rho_{l_i},\rho_{r_i})
%\end{align}
% \citep{teh2008bayesian} introduces three proposal distributions from Eq.\eqref{PostPost::Proposal}, SMC-Prior, SMC-PriorPost, and SMC-PostPost. In particular, PostPost particles are of the highest quality, because it samples exactly from the "local posterior". However, it is computationally intensive with complexity $O(n^3)$ per particle due to the pair-wise integration at each coalescent event.
%
%To address the scaling problem of SMC-PostPost, \citep{gšrŸr2009efficient} formulate coalescent as a regenerative race process. Their SMC1 algorithm does pair-wise sampling only once, and thereby reduces the complexity to $O(n^2)$ per particle. \citep{gšrŸrscalable} makes direct improvement to PostPost, by doing computations only for similar pairs. Their algorithm SMCnn further improves the computational efficiency.
SMC approaches recently published in machine learning  \citep{teh2008bayesian, gšrŸr2009efficient, gšrŸrscalable} use a similar state space representation to the MCMC approaches. They start from the current population, and iteratively select coalescent pairs and times based upon the so-called "local posterior'' distribution. \cite{teh2008bayesian} introduces three proposal distributions from the local posterior, SMC-Prior, SMC-PriorPost, and SMC-PostPost. In particular, PostPost particles are of the highest quality, because it samples exactly from the local posterior. However, it is computationally intensive with complexity $O(n^3)$ per particle due to the pair-wise integration at each coalescent event. The scalability problem is addressed in \citep{gšrŸr2009efficient} and \citep{gšrŸrscalable}. The general form of the likelihood approximation for SMC takes the following form
\begin{align}
\label{lhsurf::SMC}
L(\theta) \approx \sum_{i=1}^M w^{(i)} \frac{p_\theta(X,\mathcal{G}^{(i)})}{q_{\theta_0}(\mathcal{G}^{(i)}|X)},
\end{align}
where $w_{(i)}$ is the particle weights.

\subsection{Particle MCMC}
PMCMC \citep{andrieu2010particle} can bee viewed as a integration of MCMC and SMC (or IS) methods, where each MCMC iteration samples SMC particles to build the proposal distribution. As the SMC particles are an empirical approximation of the posterior distribution, the proposal move can be more effective, and less apt to being stuck in local minima. This methodology has not been applied to coalescent parameter inference before. We detail our contribution in this aspect in the next sections.

\section{Towards a New Inference Algorithm}
We now propose a novel approach to sample from the posterior $p_{\theta_0}(\mathcal{S}, \mathcal{T}|X)$ based upon the Particle Markov Chain Monte Carlo method. The core idea is to decouple tree structure and coalescent times, and alternately sample each of them conditioned on the other, via the Particle Gibbs Sampler (PGS) \citep{andrieu2010particle}. We hope this treatment will build proposal distributions which are close to $p(\mathcal{S},\mathcal{T}|X,\theta_0)$, and makes the sampling efficient in terms of particle quality.

First, we use Gibbs sampling to sample coalescent times from $p(\mathcal{T}|X,\mathcal{S}^k)$. Posterior probability of $t_i$ conditioned on the rest of the configuration of the genealogy is,
\begin{align}
\label{t_i::Gibbs}
&p(t_i|X, \mathcal{S}, \mathcal{T}_{-i}) \propto p(X|\mathcal{S},\mathcal{T}_{-i},t_i) p(\mathcal{S}, \mathcal{T}_{-i}, t_i)
\end{align}
The right hand side of Equation \eqref{t_i::Gibbs} is a product of \eqref{coalescent:prior} and \eqref{XcG:internal} or \eqref{XcG:root}, which only depends on $t_i$ at a few terms. More specifically, for non-root hidden notes,
\begin{align}
\label{t_i::Gibbs::internal}
 p(t_i|X, \mathcal{S}, \mathcal{T}_{-i}) \propto & \exp\left\{ -\begin{pmatrix}n-i \\2\end{pmatrix} (t_i-t_{i+1})\right\}  \cdot \exp\left\{ -\begin{pmatrix}n-i+1 \\2\end{pmatrix} (t_{i-1}-t_i)\right\} \nonumber\\
& \cdot \sum_{y_i} \sum_{y_{o(i)}} T_{o(i),i} \mb{m}_{o(i)\rightarrow i} \prod_{b=l,r} \sum_{y_{c_b(i)}} T_{i,c_b(i)} \mb{m}_{c_b(i)\rightarrow i},
\end{align}
where $\max(t_{i+1},t_{o(i)}) \leq t_i \leq \min(t_{i-1},t_{c_l(i)},t_{c_r(i)})$.
For root note,
\begin{align}
\label{t_i::Gibbs::root}
p(t_i|X, S, T_{-i}) \propto &\exp\left\{ -\begin{pmatrix}n-i+1 \\2\end{pmatrix} (t_{i-1}-t_i)\right\} \cdot \sum_{y_i}p_0(y_i) \prod_{b=l,r} \sum_{y_{c_b(i)}} T_{i,c_b(i)} \mb{m}_{c_b(i)\rightarrow i},
\end{align}
where $-\infty \leq t_i \leq \min(t_{i-1},t_{c_l(i)},t_{c_r(i)})$. $\{t_i, i=1,2,...,n-1\}$ are sampled from above equations iteratively. Note that at each iteration, the node with updated time should propagate messages outward to the rest of the whole tree, so that other nodes collect updated messages when their coalescent times are to be sampled.

Second, we use conditional SMC to sample the structure from $p(\mathcal{S} |X,\mathcal{T}^{k+1})$. $S$ corresponds to an ordered list of coalescent events $(s_1,..., s_{n-1})$. We adopt the ``local posterior" idea \citep{teh2008bayesian} to sample the coalescent pair (see Eq.(10) in \citep{teh2008bayesian} for example). But the fundamental difference is, we sample the topology conditioned on the coalescent times, so avoid the costly computation of pairwise integration. Actually, as $\delta_i$ is fixed, we only have to evaluate local posteriors at the given coalescent times,
\begin{align}
\label{proposal}
q(\rho_l,\rho_r|X,t_i)  \propto  Z^u_{i\rightarrow o(i)}(\rho_l,\rho_r,t_i)
%\exp \left(-\begin{pmatrix}n-i+1\\2\end{pmatrix} \delta_i \right)
\end{align}
for all existing pairs, and sample the tree structure from a series of multinomial distributions composed of these local posteriors. The normalization constants are,
\begin{align}
\label{proposal:constant}
& w_i = \sum_{l,r} Z^u_{i\rightarrow o(i)}(\rho_l,\rho_r,t_i), i\leq n-2 \nonumber\\
& w_{n-1} = \sum_y p_0(y) \prod_{b=l,r}\sum_{y_{c_b}} T_{2n-1,c_b(2n-1)}  \mb{m}_{c_b(2n-1)\rightarrow 2n-1}
\end{align}
Suppose SMC generates $M$ particles, then the posterior is approximated by,
\begin{align}
\hat{p}(S|X,T) = \sum_{m=1}^M W^{(m)} \delta_{S^{(m)}}(S),
\end{align}
where,
\begin{align}
W^{(m)} &\propto \frac{p(S^{(m)},T,X)}{q(S^{(m)}|X,T)} \propto \frac{p(X|S^{(m)},T)}{q(S^{(m)}|X,T)}
\end{align}
The second ratio holds because $p(\mathcal{S}^{(m)},\mathcal{T})$ is constant regardless of $\mathcal{S}^{(m)}$ by Eq.\eqref{coalescent:prior}. Moreover, the numerator $p(X|\mathcal{S}^{(m)},\mathcal{T})$ is computed with Eq.\eqref{XcG:root}.  For Sequential Importance Sampling (SIS) (a special case of SMC without resampling), the weights are,
\begin{align}
\label{weight::SIS}
W^{(m)} \propto \prod_{i=1}^{n-1} w_i^{(m)}
\end{align}
using Equation \eqref{XcG:root}, \eqref{proposal} and \eqref{proposal:constant}. For SMC described by \cite{andrieu2010particle}, the resampling step at each iteration $i$ implicitly counts for the weight $\mathbf{w} _{i-1}$, so the particle weight is just the weight of the last iteration, i.e.,
\begin{align}
\label{weight::SMC}
W^{(m)} & \propto \frac{p(S^{(m)}_{1:n-1},T,X)}{p(S^{(m)}_{1:n-2},T,X) q(S^{(m)}_{n-1}|S^{(m)}_{1:n-2})}  \propto w_{n-1}^{(m)}
\end{align}
To this point, we specify our SMC procedure to build a particle approximation of the conditional distribution $\hat{p}(\mathcal{S}|X,\mathcal{T})$. As noted by \cite{andrieu2010particle},  we keep the structure from previous PGS iteration as one current particle to follow the legitimate cSMC requirement. Finally, we sample the new structure $S^{k+1}$ from the set of weighted cSMC particles.

Our Particle Gibbs Sampling algorithm for Kingmans' n-Coalescent (PGS-Coalescent) is described in Algorithm \ref{Alg::PGS::Coalescent}. It computes the relative likelihood of $\theta$ as,
\begin{align}
\label{lhsurf::PGS}
\frac{L(\theta)}{L(\theta_0)} \approx \frac{1}{M} \sum_{i=1}^M \frac{p(X|\mathcal{G}^{(i)}, \theta)}{p(X|\mathcal{G}^{(i)}, \theta_0)}
\end{align}
\begin{algorithm}[htp]
\caption{Particle Gibbs Sampler for the Kingman's Coalescent (PGS-Coalescent, abbr. PGSC)}
\label{Alg::PGS::Coalescent}
\begin{algorithmic}
\STATE Initialize $(\mathcal{S}, \mathcal{T})=(\mathcal{S}^0, \mathcal{T}^0)$ randomly, where $\mathcal{S}$ is genealogy structure, and $\mathcal{T}={t_1,...,t_{n-1}}$ is time of coalescent events.
\REPEAT
\STATE 1) sample $\mathcal{T}^{k+1}\sim p(\mathcal{T}|X,\mathcal{S}^k)$ using Gibbs sampling with Equation \eqref{t_i::Gibbs::internal}, \eqref{t_i::Gibbs::root}, update messages with Equation \eqref{msg:internal}, \eqref{msg:root} after each $t_i$ is changed
\STATE 2) run a conditional SMC algorithm targeting $p(\mathcal{S}|X,\mathcal{T}^{k+1})$ with proposal distribution \eqref{proposal}, and get a set of weighted particles $\{\mathcal{S}^\ast_1,...,\mathcal{S}^\ast_N\}$ which include $\mathcal{S}^k$
\STATE 3) sample $\mathcal{S}^{k+1}\sim \hat{p}(\mathcal{S}|X,\mathcal{T}^{k+1})$ with Equation \eqref{weight::SIS} or \eqref{weight::SMC}
\UNTIL{Equilibrium}
\end{algorithmic}
\end{algorithm}

\section{Applications}
We compare the performance of PGS-Coalescent with the IS, MCMC and SMC based methods in this section. We simulate coalescent samples and use Equation \eqref{lhsurf::IS}, \eqref{lhsurf::MCMC}, \eqref{lhsurf::SMC} and \eqref{lhsurf::PGS} to evaluate the likelihood surface of $\theta$. Our major concern is particles' quality. Specifically, good approaches should use fewer samples to reconstruct more accurate likelihood surfaces. We also monitor CPU time, but only for the purpose of evaluating scalability, as different algorithms are implemented in various languages. Experiments are performed on an Intel i3 PC. To make the time scalability analysis reliable, we run each algorithm without any parallelization.

\subsection{Binary Alleles}
The dataset is from \citep{griffiths1994simulating}. It has 50 individuals of 20 sites. Each site has binary alleles. The unit mutation rate matrix is,
\begin{align}
R = \begin{pmatrix} -0.5 & 0.5 \\ 0.5 & -0.5 \end{pmatrix}
\end{align}
We run IS, MCMC, SMC1 and PGSC on this dataset randomly for five times. The (relative) likelihood surface is shown in Figure \ref{sq}. We can see MCMC method and SMC method performs not good in this dataset. The MCMC method didn't capture the mode of likelihood surface. The SMC method can get a rough shape of the surface, but the magnitude varies a lot. PGSC generate comparable results to IS, the state-of-the-art, with much fewer samples.
\begin{figure}[!ht]
\centering
\subfigure[]{
\includegraphics[width=2.8in]{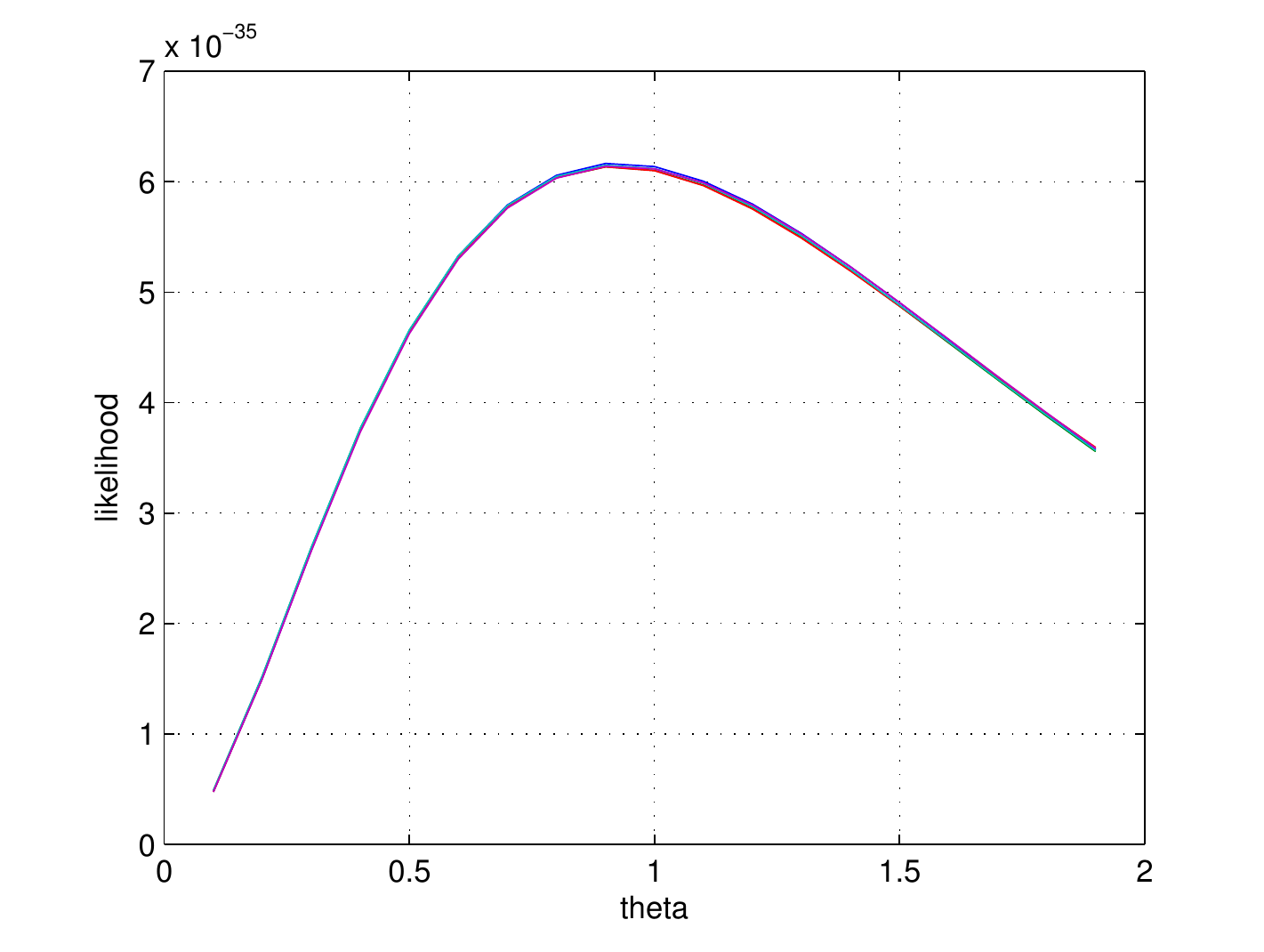}
\label{sq:: IS}
}
\subfigure[]{
\includegraphics[width=2.8in]{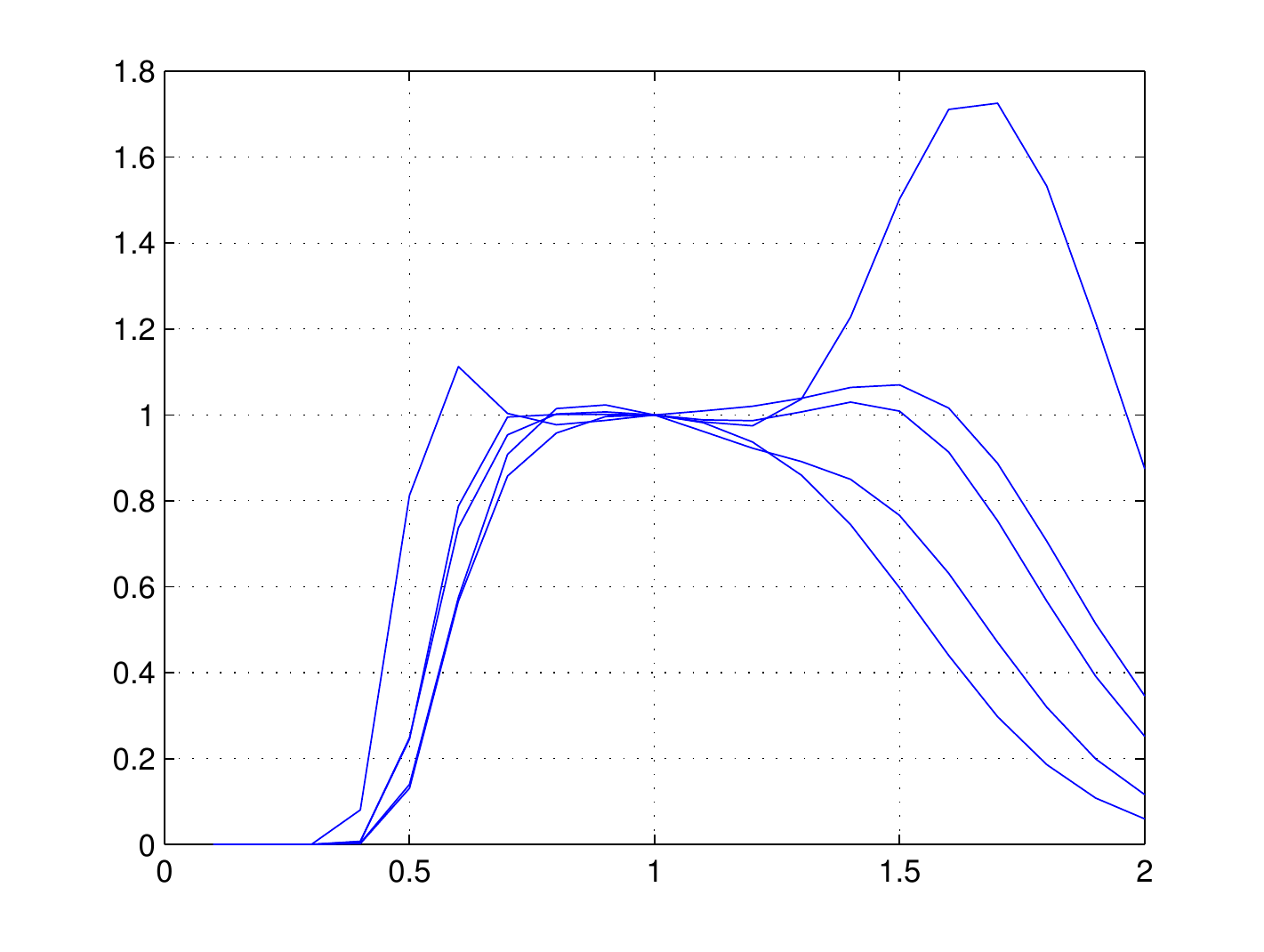}
\label{sq::MCMC}
}
\subfigure[]{
\includegraphics[width=2.8in]{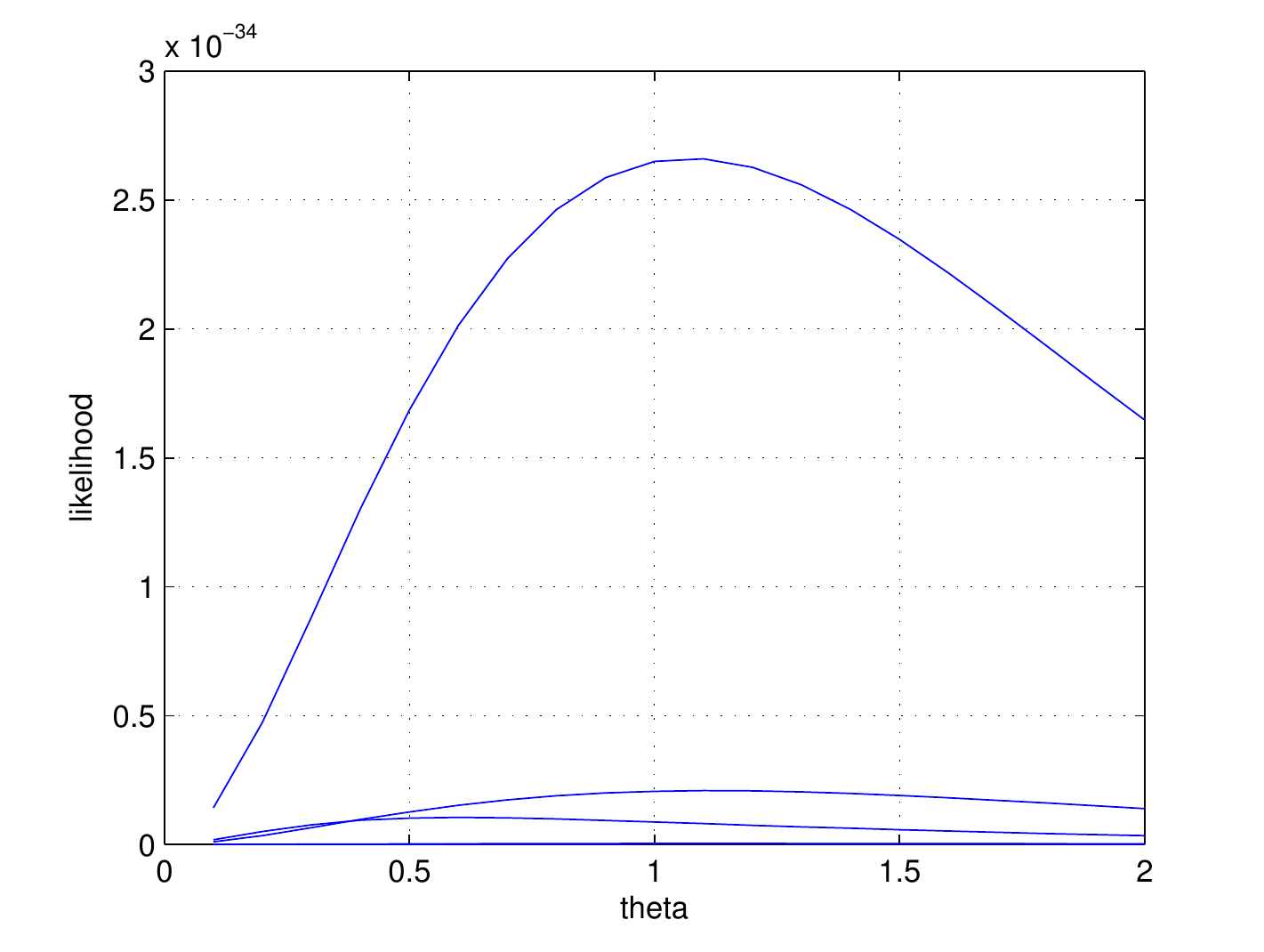}
\label{sq::SMC1}
}
\subfigure[]{
\includegraphics[width=2.8in]{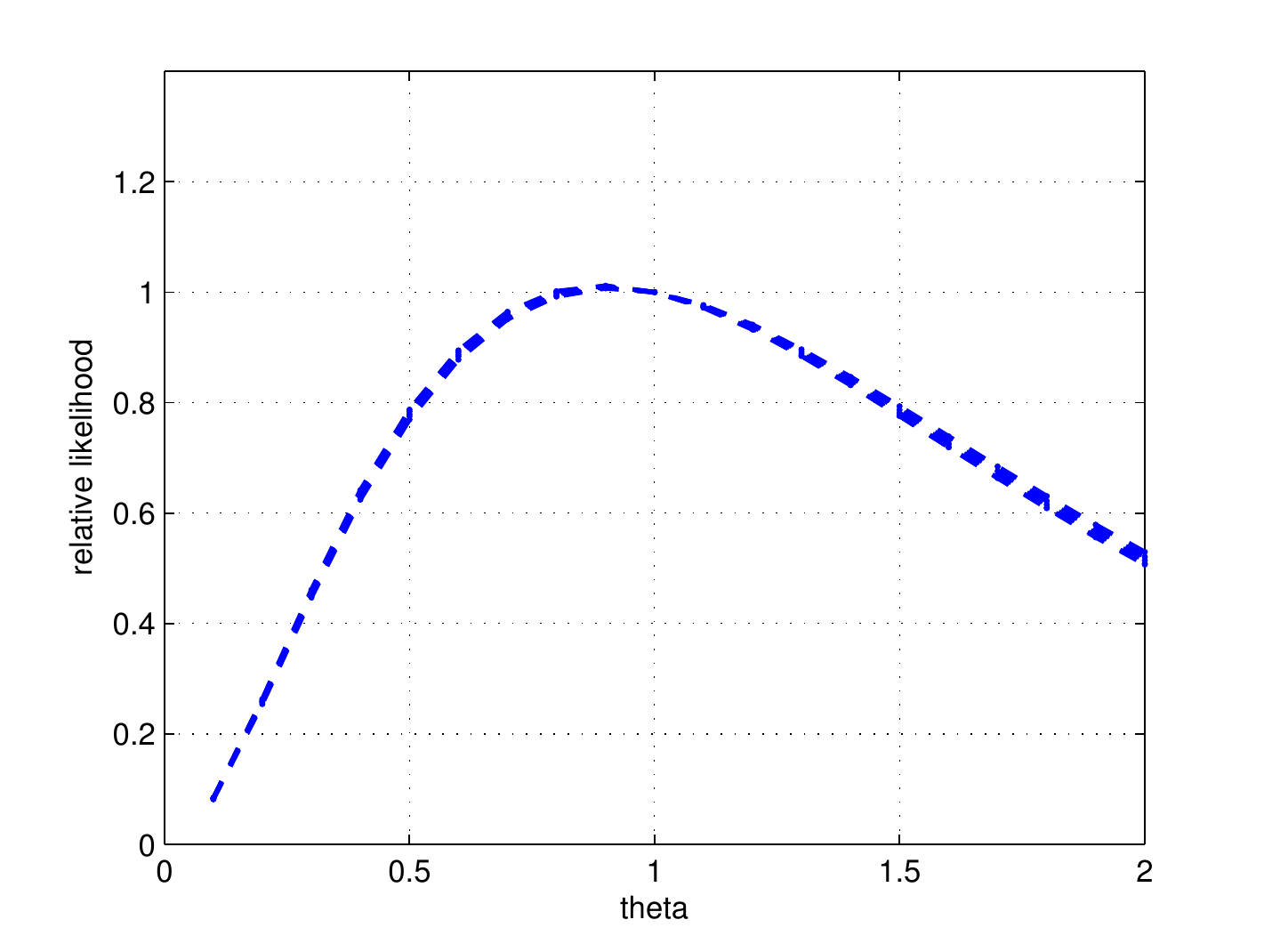}
\label{sq::PGSC}
}
\label{sq}
\caption[Optional caption for list of figures]{(Relative) likelihood surface of sequence data from \citep{griffiths1994simulating} by different methods, each run randomly for 5 times. (a) 10000 samples of IS method; (b) 1,000,000 iterations of MCMC method, first 50,000 iterations discarded, the rest thinned at interval of 200, so 4750 valid samples; (c) 50000 samples of SMC1 method; (d) 2000 iterations of PGSC, each with 200 cSMC particles and 50 Gibbs sampling rounds, discard first 800 iterations, no thinning, yielding 1200 valid samples \label{sq}}
\end{figure}

\subsection{Microsattelites}
Next, we consider the microsatellite data used by \cite{stephens2000inference}, with state space $\{0,1,...,19\}$, under stepwise mutation model, where each state can only mutate to one of its neighbors with equal probability.

\subsubsection{One-locus case}
The data is $\{8,11,11,11,11,12,12,12,12,13\}$. We run IS, MCMC, SMC1 and PGSC on it randomly for five times. The (relative) likelihood surface is shown in Figure \ref{ms1}. In this simple scenario of one-site and ten-individual, IS, SMC1 and PGSC methods perform well, while SMC1 is slightly better than the other two. MCMC methods still cannot capture the accurate shape of the likelihood surface. Again we note that PGSC uses fewest samples to achieve comparable results to SMC1 and IS.
\begin{figure}[!ht]
\centering
\subfigure[]{
\includegraphics[width=2.8in]{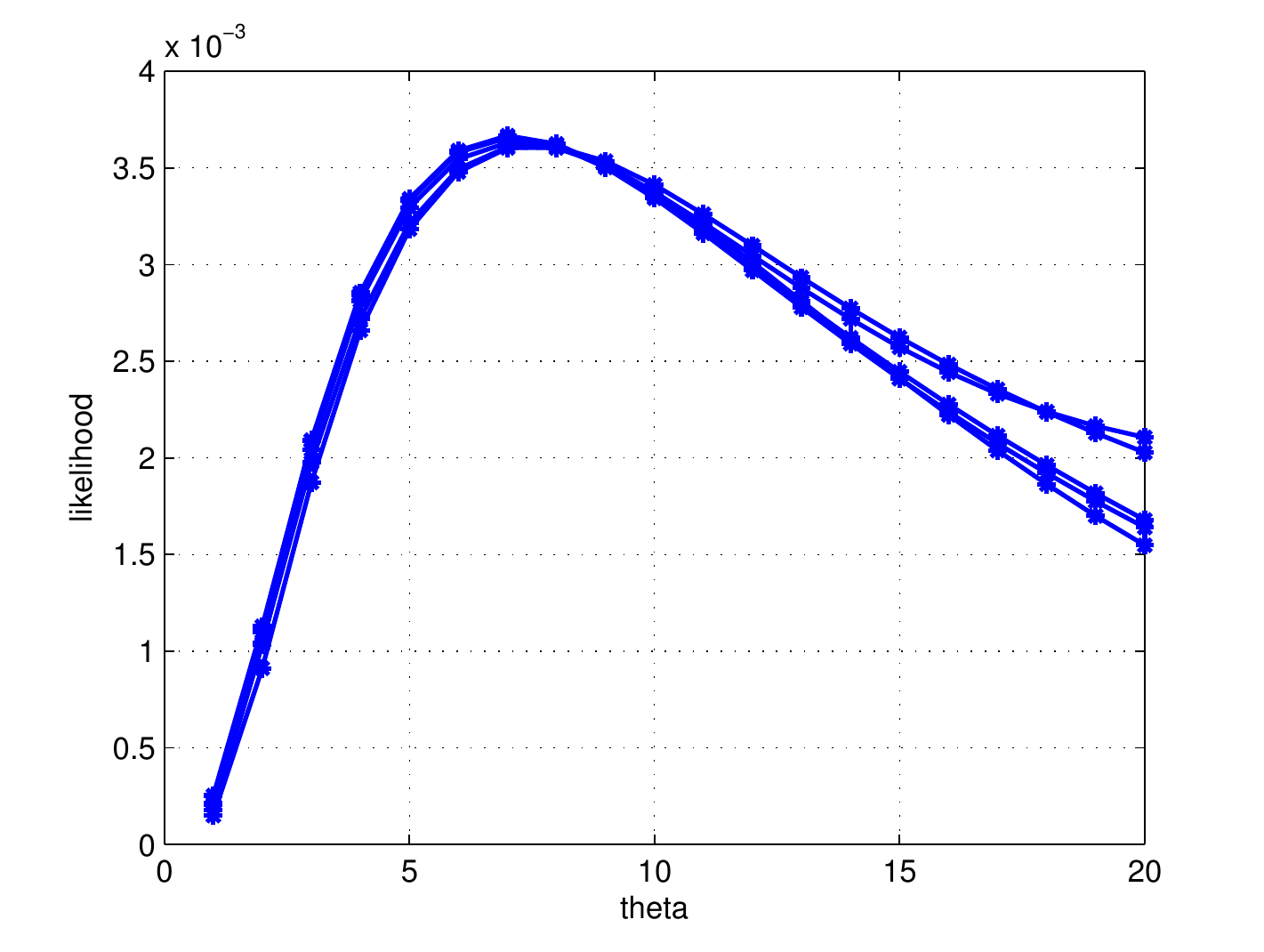}
\label{ms1:: IS}
}
\subfigure[]{
\includegraphics[width=2.8in]{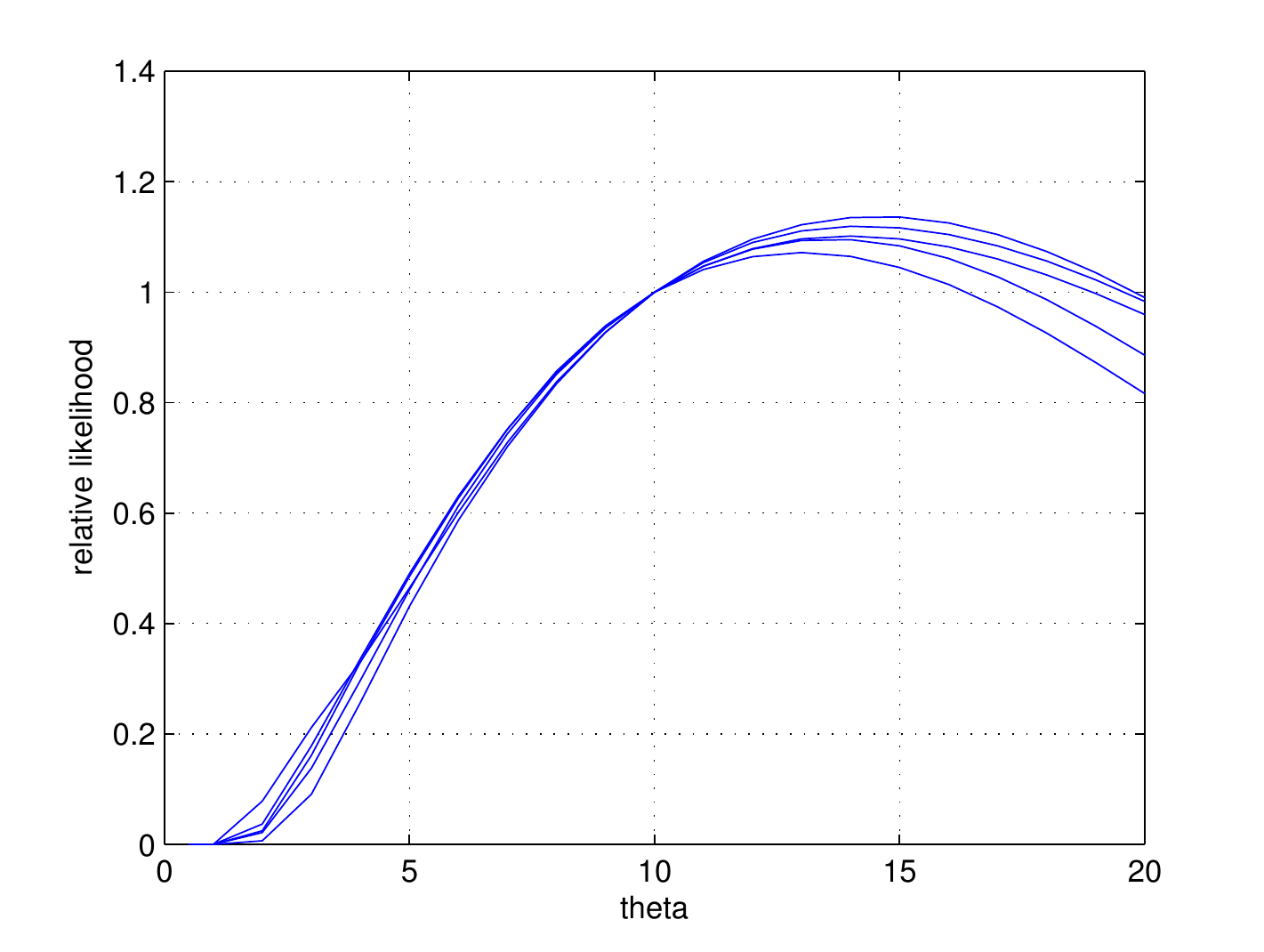}
\label{ms1::MCMC}
}
\subfigure[]{
\includegraphics[width=2.8in]{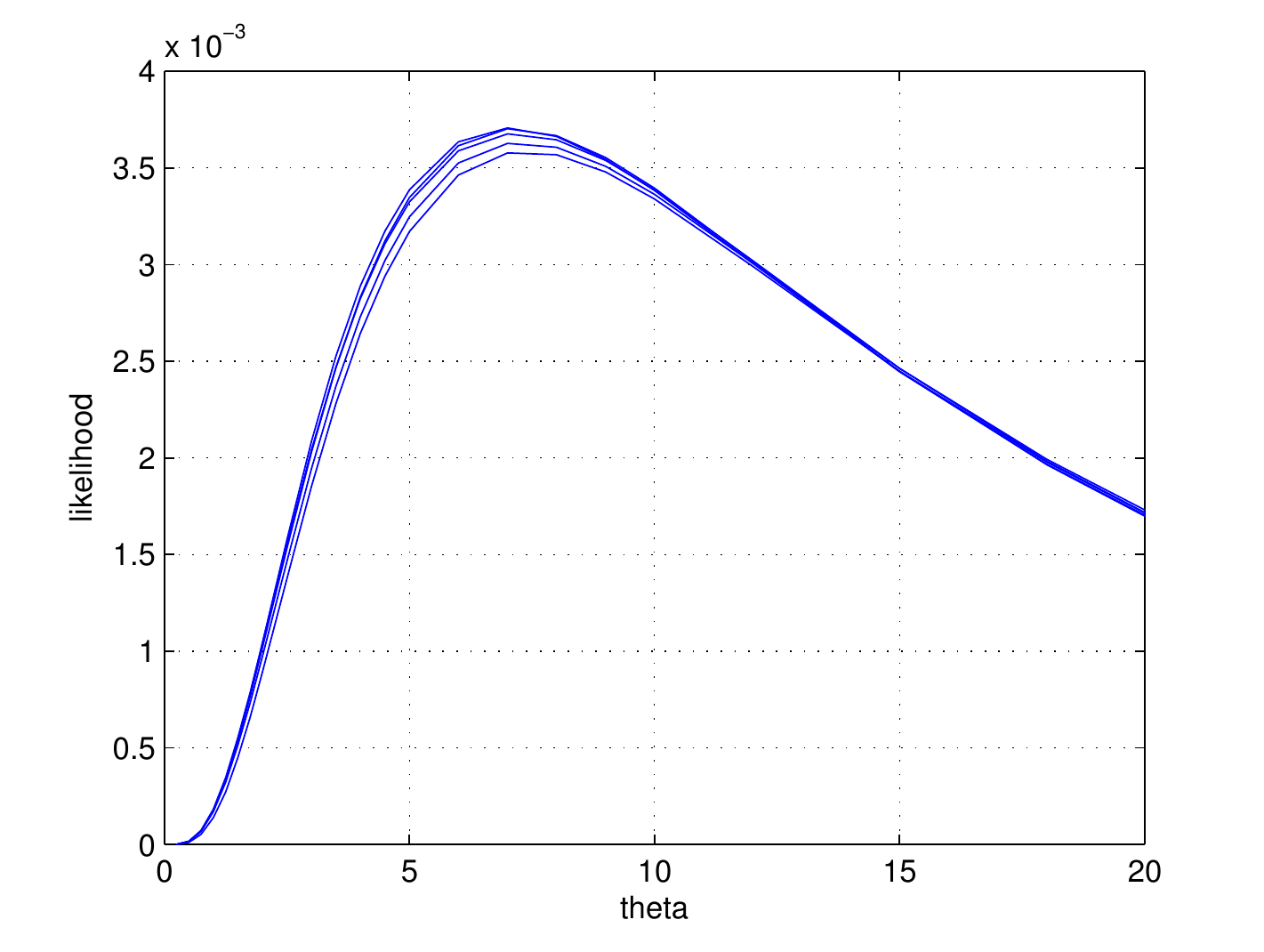}
\label{ms1::SMC1}
}
\subfigure[]{
\includegraphics[width=2.8in]{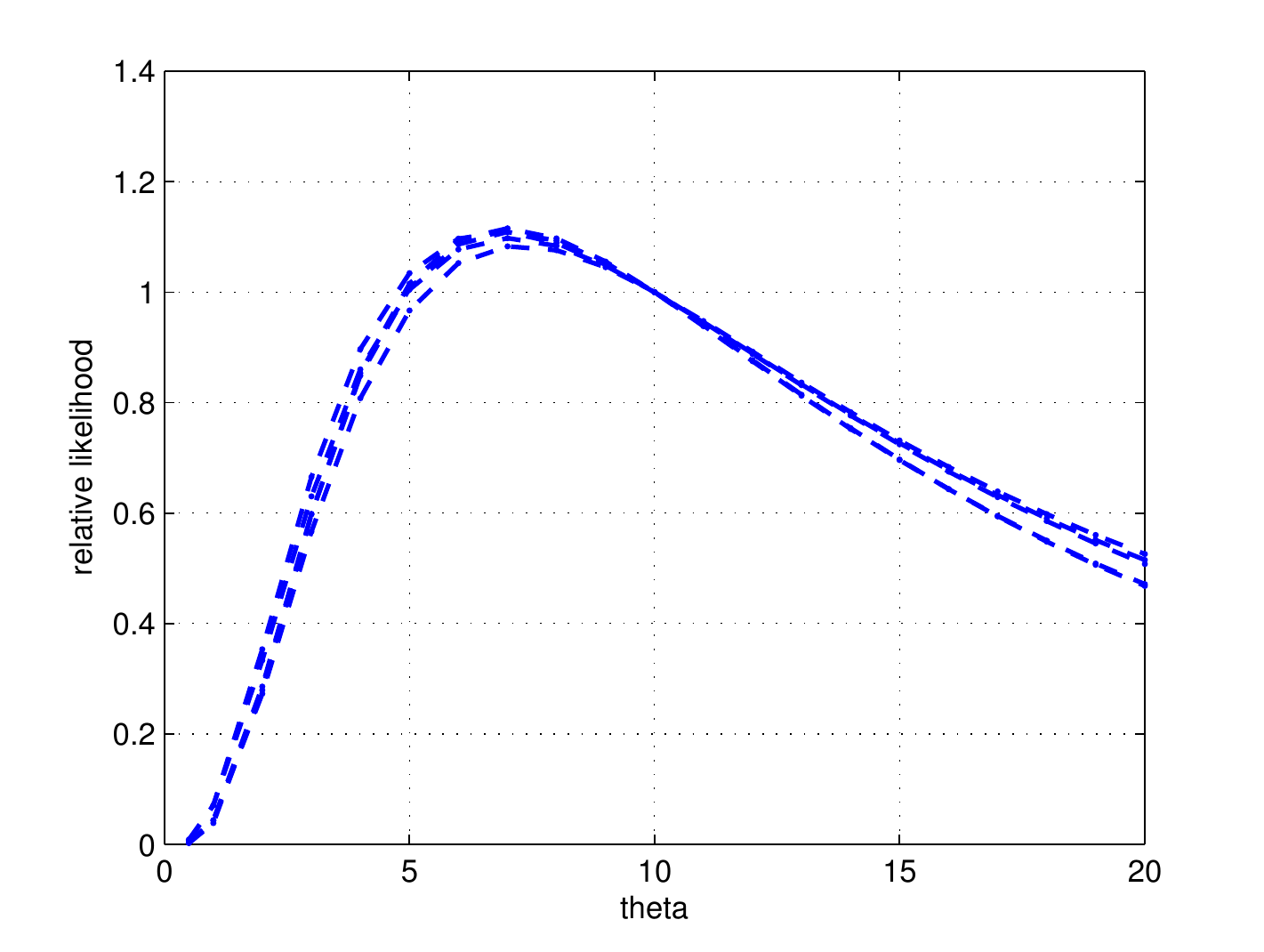}
\label{ms1::PGSC}
}
\label{ms1}
\caption[Optional caption for list of figures]{(Relative) likelihood surface of one-locus microsatellite data from \citep{stephens2000inference} by different methods, each run randomly for 5 times. (a) 10000 samples of IS method; (b) 1,000,000 iterations of MCMC method, first 50,000 iterations discarded, the rest thinned at interval of 200, so 4750 valid samples; (c) 10000 samples of SMC1 method; (d) 1000 iterations of PGSC, each with 40 cSMC particles and 10 Gibbs sampling rounds, discard first 400 iterations, no thinning, yielding 600 valid samples \label{ms1}}
\end{figure}

\subsubsection{Five-loci case}
The dataset consists of 60 5-sites individuals. We run IS, MCMC, SMC1 and PGSC on it randomly for four times. The (relative) likelihood surface is shown in Figure \ref{ms5}. This is a challenging example, and all of these methods show some drawbacks. The IS method can capture the shape and mode well but not the magnitude, while SMC1 only get a rough shape. As before, MCMC still cannot capture the mode. PGSC is performing comparable to IS, but the diversity of particles becomes poor.
\begin{figure}[!ht]
\centering
\subfigure[]{
\includegraphics[width=2.8in]{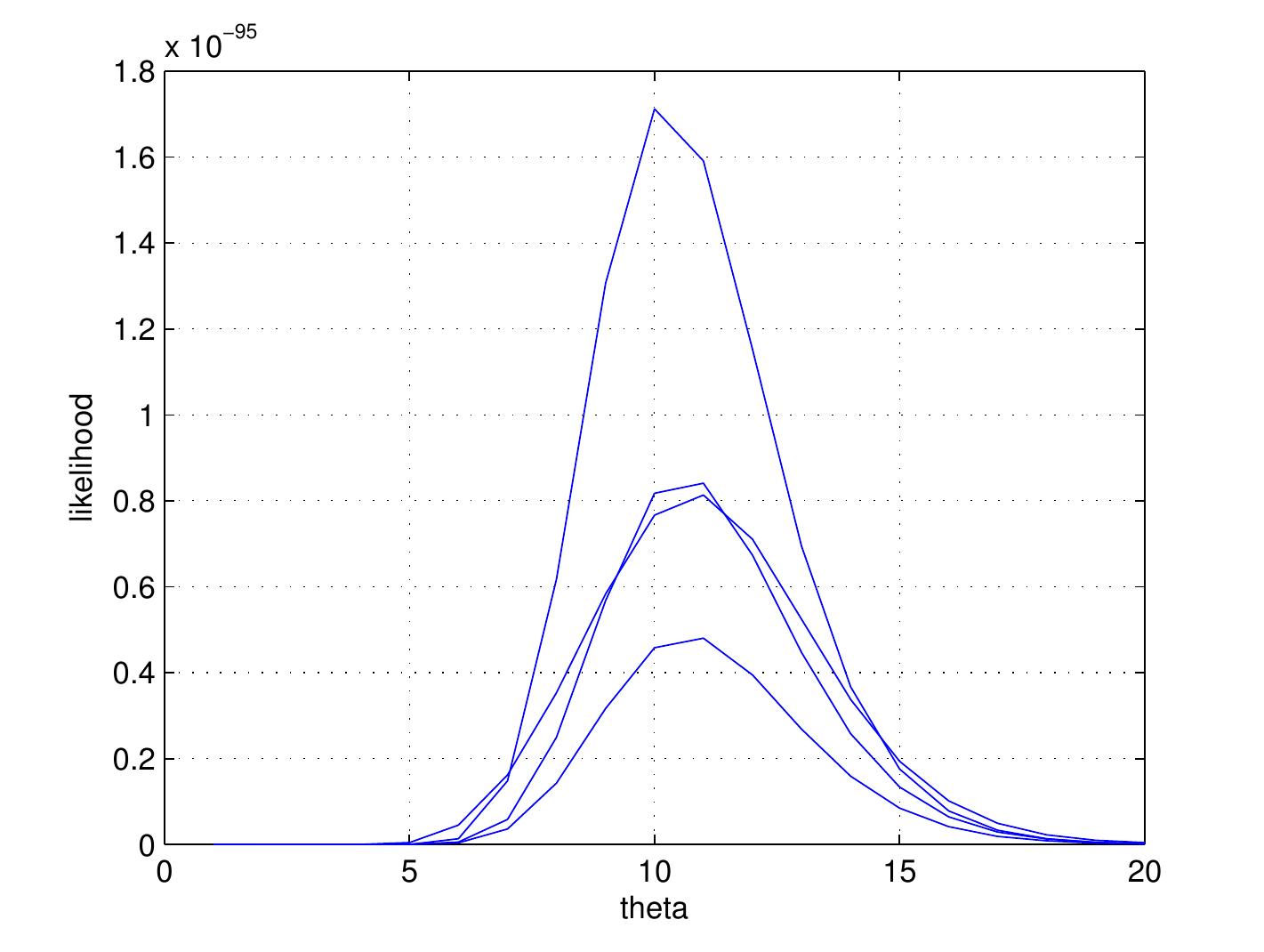}
\label{ms5:: IS}
}
\subfigure[]{
\includegraphics[width=2.8in]{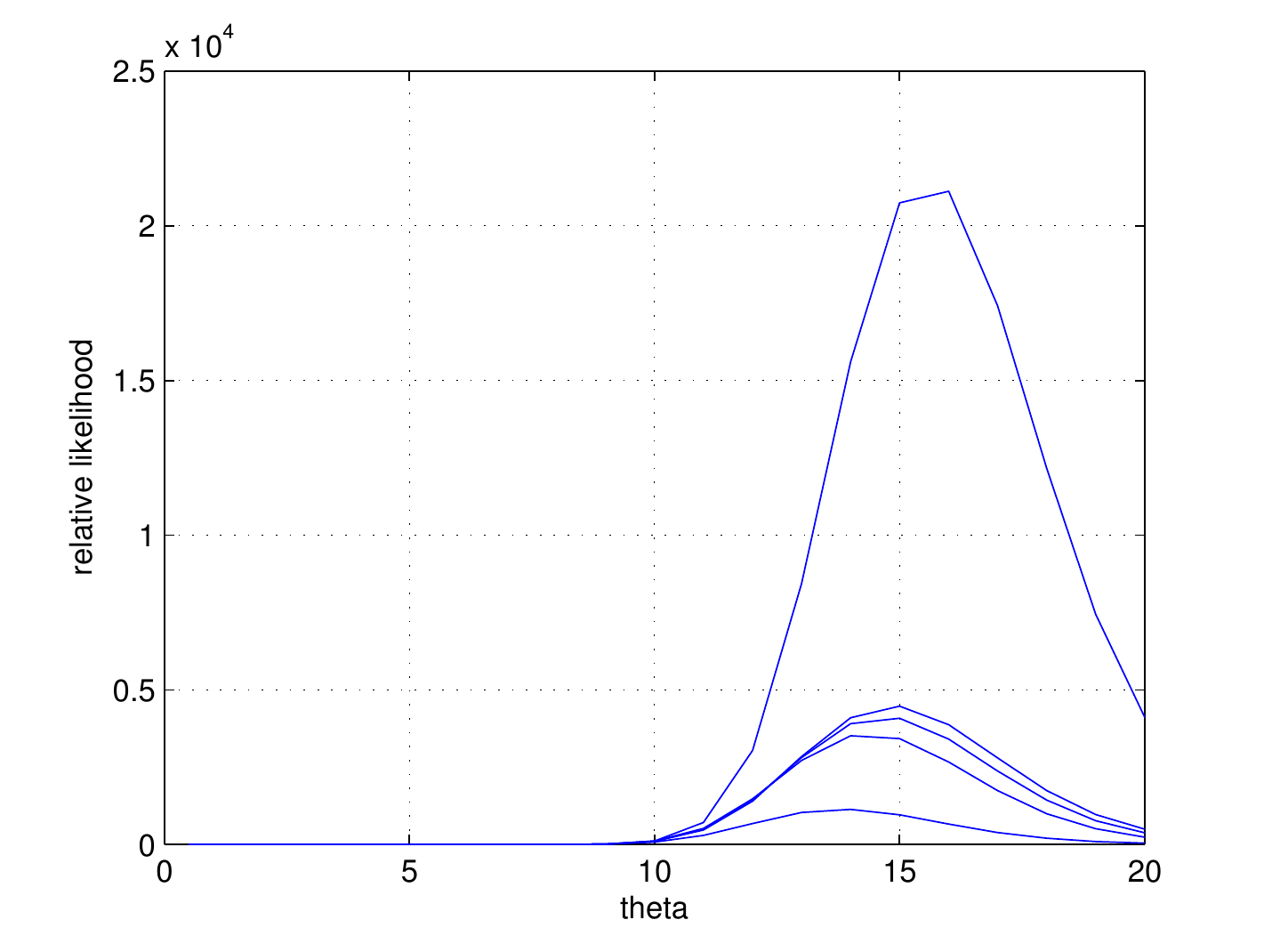}
\label{ms5::MCMC}
}
\subfigure[]{
\includegraphics[width=2.8in]{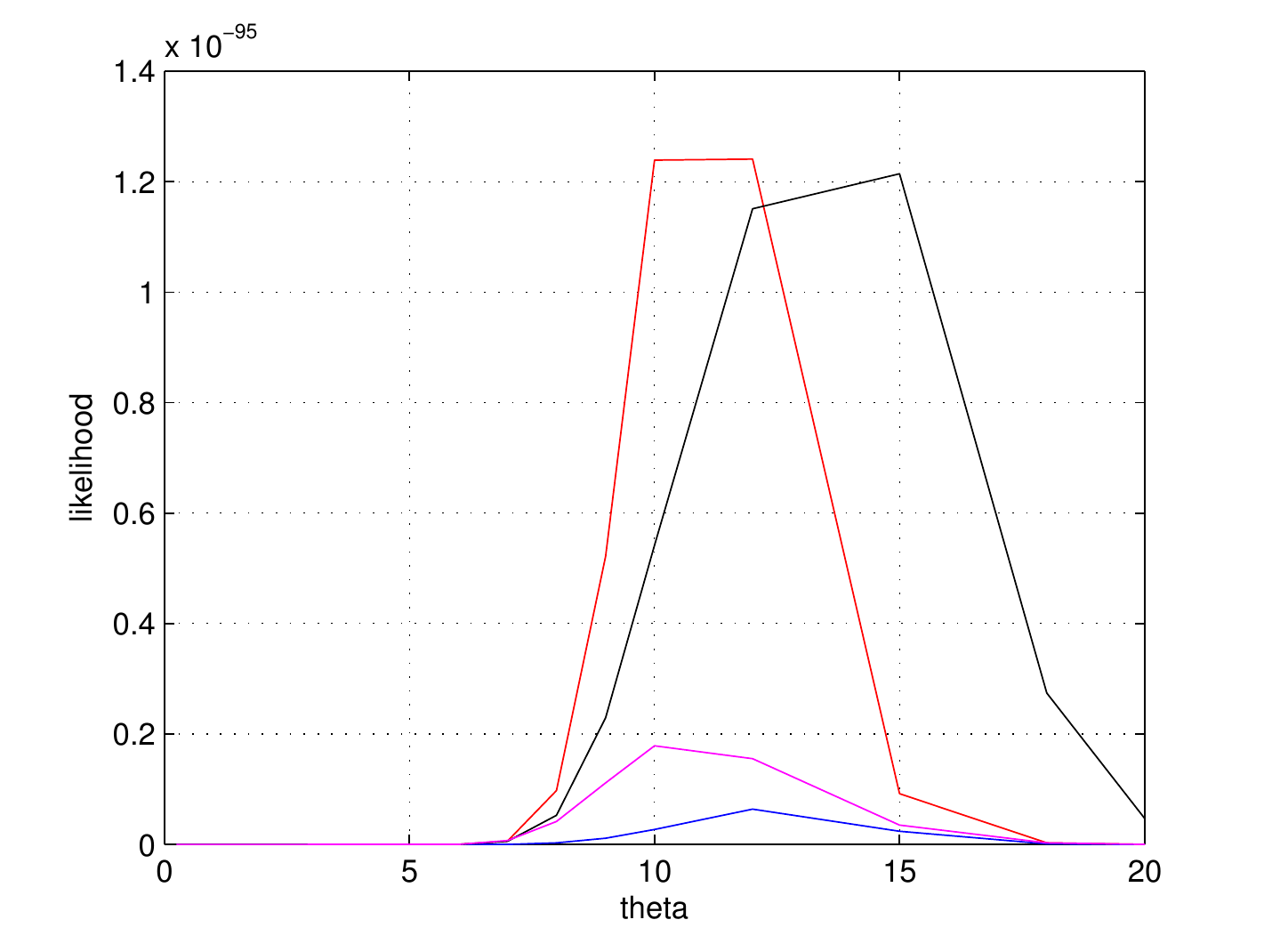}
\label{ms5::SMC1}
}
\subfigure[]{
\includegraphics[width=2.8in]{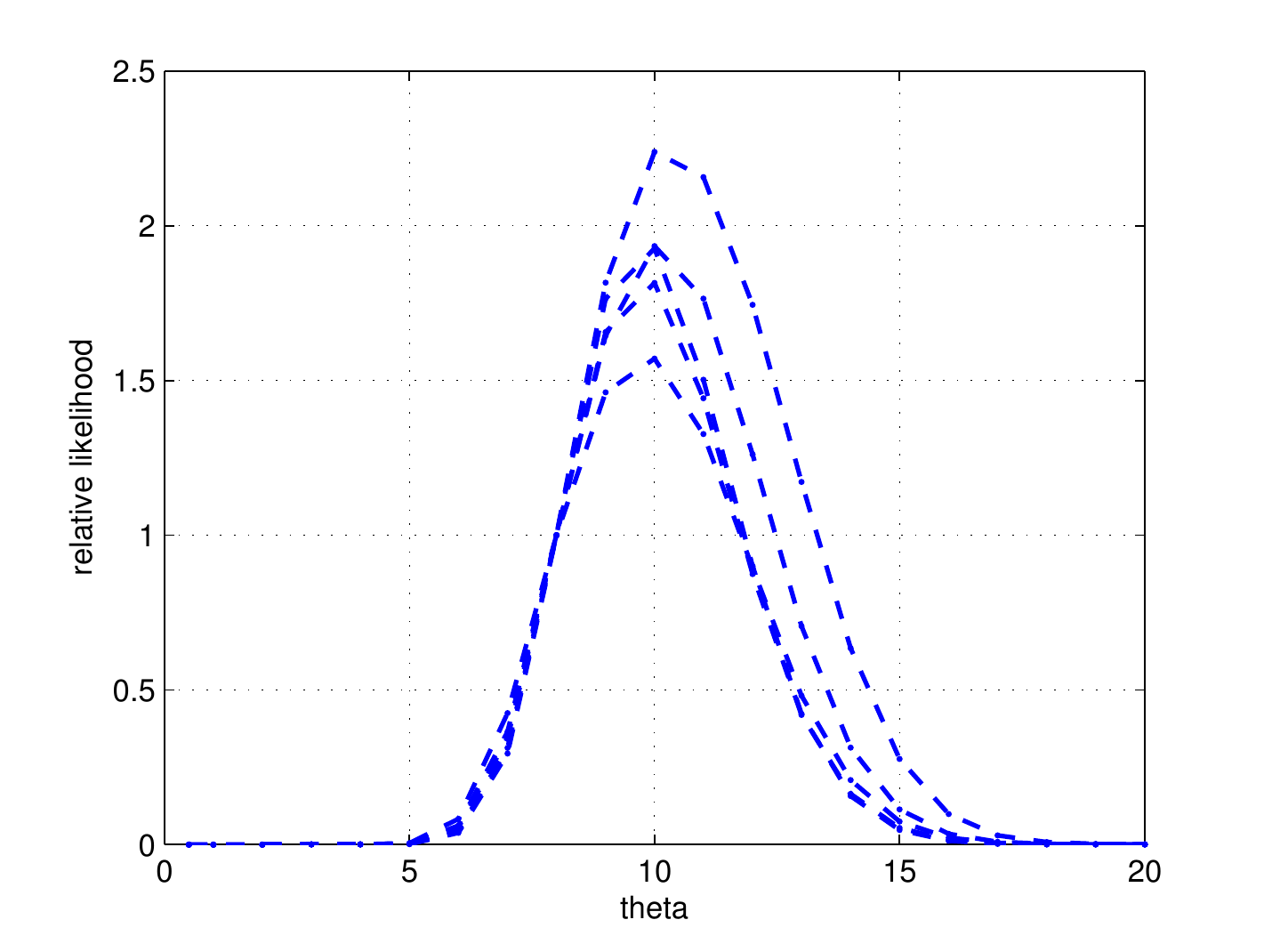}
\label{ms5::PGSC}
}
\label{ms5}
\caption[Optional caption for list of figures]{(Relative) likelihood surface of five-locus micro-satellite data from \citep{stephens2000inference} by different methods, each run randomly for 4 times. (a) 500,000 samples of IS method; (b) 1,000,000 iterations of MCMC method, first 50,000 iterations discarded, the rest thinned at interval of 200, so 4750 valid samples; (c) 10000 samples of SMC1 method; (d) 2000 iterations of PGSC, each with 200 cSMC particles and 60 Gibbs sampling rounds, discard first 1200 iterations, no thinning, yielding 160 valid samples \label{ms5}}
\end{figure}

\subsection{Scalability evaluation}
In previous experiments, the problem scale increases from microsatellite one-locus data, to binary allele data, and then microsatellite five-locus data. The CPU time to generate results shown in Figure \ref{sq}, \ref{ms1} and \ref{ms5} is recorded in Table \ref{expTab} (averaged over the several random runs for each method). We can see the MCMC method is least sensitive to problem scale. This is expected because its proposal distribution is simplest among all the methods. Comparing the two best approches in terms of accuracy, i.e., IS and PGSC, one can see IS is much faster than PGSC in all cases. However, IS slows down quickly when the data size and diversity grows up, and PGSC looks better in terms of scalability.
\begin{table}[!ht]
	\centering
	\caption{CPU time (in seconds) of different algorithms on previous experiments}
	\captionsetup{belowskip=0.5pt}
	\label{expTab}
	\begin{tabular}{ | l | l | l | l | }
		\hline
		Method & One-locus microsatellite data & Binary allele data & Five-loci microsatellite data \\
 		\hline
		IS  &    8  &  65  &  15300 \\
    		MCMC  &  13450  &  56000   &  76100  \\
       		SMC1 &  1693.3 &   NA  &  43200  \\
		PGSC & 905 & 13050 & 71000 \\
  		\hline
	\end{tabular}
\end{table}

\section{Discussion and Conclusion}
Algorithm \ref{Alg::PGS::Coalescent} includes Gibbs Sampling and cSMC as two sub-routines during each iteration. Rigorously, its complexity depends on three factors. 1) $m_\mathcal{T}$, number of iterations to reach equilibrium of $p(\mathcal{T}|X,\mathcal{S})$; 2) $m_\mathcal{S}$, particle capacity that moderately models $p(\mathcal{S}|X,\mathcal{T})$; 3) $m_\mathcal{G}$, number of iterations of PGS to reach equilibrium of $p(\mathcal{G}|X)$. Taking BP and structural sampling of Equation \eqref{proposal} into consideration, the complexity is $O(m_\mathcal{G}(m_\mathcal{T}n^2+m_\mathcal{S}n^3))$. Although computationally intensive, we observed that the algorithm performs reasonably efficient in practice, thanks to two advantages. First, cSMC provides a good approximation of $p(\mathcal{S}|X,\mathcal{T})$. Second, $m_\mathcal{T}, m_\mathcal{S}$ can be set small but still generate high quality samples. These can be seen from previous experimental performance.

In essential, PGSC is an MCMC approach, but our algorithm is fundamentally different from the Metropolis-Hastings based method \citep{kuhner1995estimating, wakeley2009coalescent}, where structures and times are updated independently, and the rearrangement of the tree structure is random. In terms of state space representation, we are very close to the SMC approaches \citep{teh2008bayesian, gšrŸr2009efficient, gšrŸrscalable}. However, SMC is a sub-routine in our framework, and the conditional sampling $\mathcal{S}|\mathcal{T}$ avoids intensive pair-wise integration, which the previous works have to deal with. Our approach is also fundamentally different to the IS method \citep{stephens2000inference} in both state-space representation and methodology. The IS method explicitly simulates each mutation event, and coalescent events only happen between identical individuals. This could bring computational trouble when the data is of high dimension and very diverse. We already observe this when studying the five-locus microsatellite dataset.

To conclude, this paper has introduced a novel inference algorithm (PGS-Coalescent) for Kingman's n-Coalescent based upon the Particle MCMC methodology. It alternately samples tree structures and coalescent times conditioned on one another. We illustrate the utility of our algorithm by a parameter estimation task in population genetics ($\theta$). Experimental results show that the proposed method performs comparable to or better than several well-developed approaches. PGS-Coalescent could be further optimized by designing other efficient proposal distributions and reusing particles, which will help to improve conditional SMC and to reduce computational intensity.

\clearpage

%\subsubsection*{Acknowledgements}
%This work is supported by ... grant. We thank ... for advices.

%\subsubsection*{References}

%\bibliographystyle{plain}
\bibliographystyle{abbrvnat}
\bibliography{myref}

\end{document}